\documentclass{llncs}

\usepackage[english]{babel}
\usepackage[latin1]{inputenc}
\usepackage{graphicx}
\usepackage{epstopdf}
\usepackage{verbatim}
\usepackage{amsmath}
\usepackage{amssymb}
\usepackage{enumerate}
\usepackage{color}
\usepackage{soul}
\usepackage{cite}
\usepackage{float}
\usepackage{wrapfig}

\usepackage{fancyhdr}
\usepackage{amssymb,amsmath}
\usepackage{amsfonts}
\usepackage{setspace}
\usepackage{indentfirst}
\usepackage{algorithm}
\usepackage[noend]{algpseudocode}
\usepackage{adjustbox}
\usepackage{multirow}
\usepackage{supertabular}

\usepackage{array}
\usepackage{listings}

\title{\LARGE \bf
AIM in Osteoporosis
}

\author{Sokratis Makrogiannis \and Keni Zheng}
\institute{Division of Physics, Engineering, Mathematics and Computer Science, Delaware State University, Dover, DE 19901-2277\\ \email{smakrogiannis@desu.edu}
}

\begin{document}

\maketitle
\thispagestyle{empty}
\pagestyle{empty}

\begin{abstract}
In this chapter we explore and evaluate methods for trabecular bone characterization and osteoporosis diagnosis with increased interest in sparse approximations. We first describe texture representation and classification techniques, patch-based methods such as Bag of Keypoints, and more recent deep neural networks. Then we introduce the concept of sparse representations for pattern recognition and we detail integrative sparse analysis methods and classifier decision fusion methods. We report cross-validation results on osteoporosis datasets of bone radiographs and compare the results produced by the different categories of methods. We conclude that advances in the AI and machine learning fields have enabled the development of methods that can be used as diagnostic tools in clinical settings.
\end{abstract}


\section{Introduction}

Osteoporosis is a skeletal disorder characterized by decreased bone strength that may lead to susceptibility of fracture \cite{Bartl2009}. There are more than 3 millions of people diagnosed with osteoporosis in the U.S. per year. The risk is increasing with age, especially the people who are over 40. Timely diagnosis of osteoporosis can effectively predict fracture risk and allow for effective treatment. 

Trabecular bone characterization and automated and accurate diagnosis of bone osteoporosis is significant for improving public health. Aerial Bone Mineral Density (BMD) is computed in dual-energy X-ray absorptiometry (DXA) scans to diagnose osteoporosis \cite{hough1998}. However, BMD can predict fracture with only 60\% accuracy. Analysis of trabecular bone microarchitecture can significantly improve the prediction rates, but this information requires bone biopsy with histomorphometric analysis. The task of obtaining trabecular bone microarchitecture information by noninvasive methods is a nontrivial scientific problem \cite{macintyre2012}. Diagnosis of osteoporosis using bone radiograph scans presents some challenges, mainly because images of osteoporotic and healthy subjects are visually very similar. Previous approaches to evaluating bone structure on  radiographs by 2D texture analysis were reported in \cite{hough1998, martin2003, yger2014}. Moreover, in \cite{jennane2007, jennane2001} the authors propose to use 2D texture analysis to characterize 3D bone microarchitecture. 

Here we present and evaluate mathematical methods and algorithms for computer aided-diagnosis. The application domain is osteoporosis diagnosis in radiographs of the calcaneus bone. We will explore the use of sparse modeling and classification for classifying diseased from healthy subjects. Then we will present ensemble sparse techniques to find more accurate solutions than individual classification techniques. We will also test other classification techniques based on texture features, or patch-based techniques such as the Bag of Keypoints and deep learning methods.

\section{Methods}

\subsection{Non-sparse Classification Techniques: Texture-based, Patch-based and Deep Learning}
\subsubsection{Introduction to Texture-based Classification}
Texture is an image property that can be used for segmenting and classifying images into different objects. We can define texture as a structure consisting of a group of related elements \cite{Sonka2014}. The pixels in this group are called texture primitives, texture elements, or texels. 

Texture analysis techniques are mainly applied to texture recognition and texture based shape analysis \cite{Sonka2014}. Generally, people consider texture as fine when the texture element is small and there are large differences between elements, and coarse when the element is large and there are only few elements in the image, grained and smooth. For scientific applications of texture, we use more precise characteristics such as tone and structure \cite{Haralick1979}. Tone is more about pixel intensity and structure is about the spatial relationship between texture elements. There are many methods for texture extraction, such as wavelet analysis, Gabor filters, co-occurrence matrices, intensity histogram-based, and spatial frequency domain descriptors. 

We present texture-based methods for computer-aided diagnosis of diseased and healthy subjects \cite{Zheng2016}. Our premise is that the deterioration of disease can be captured by textural features. We first computed texture features based on wavelet decomposition, discrete Fourier and Cosine transforms, fractal dimension, statistical co-occurrence indices, and structural texture descriptors. We employed feature selection techniques that consider the individual feature predictive ability and inter-feature redundancy to find the most discriminant feature set. In the classification stage we employed Na\"{i}ve Bayes, Multilayer Perceptron, Bayes Network, Random Forests and Bagging models for diagnosis.

\subsubsection{Feature Computation}
The purpose of this stage is to compute texture descriptors that can be used for separation between groups of healthy and diseased subjects. This is usually performed in a high-dimensional feature space to reduce the Bayes error rate. Next, we describe frequently used feature sets. 
\paragraph{Fractal Dimension}    
These features have shown promise in texture classification applications. A fractal is defined as a mathematical set whose Hausdorff dimension exceeds the fractal's topological dimension \cite{Pentland1984}. It has been shown that fractal dimension correlates well with a function's roughness. Therefore, we used fractal dimension to measure the roughness and granularity of the image intensity function. The topological dimension of this function is equal to 3, consisting of 2 spatial dimensions plus the intensity. 

\begin{figure}[t]
\centering
\includegraphics[scale=0.6]{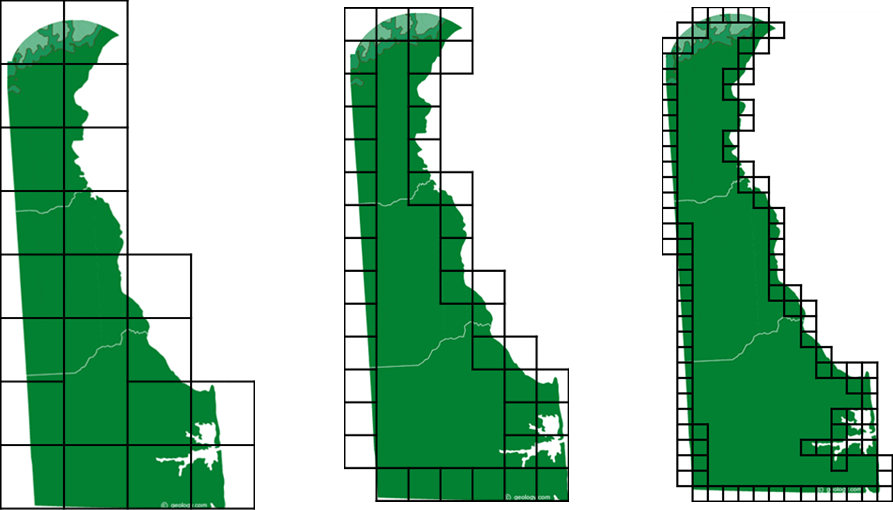}
\caption{Box counting to compute the fractal dimension of Delaware state boundary.}
\label{fig_FD}
\end{figure}

The method of box counting can be utilized to compute the fractal dimension. Assuming a fractal structure with dimension $D$, we let $N(\epsilon)$ be the number of non-empty boxes of size $\epsilon$ required to cover the fractal support. Using the relation $N(\epsilon) \simeq \epsilon ^{(-D)}$, we can numerically estimate $D$ from 
\begin{equation}
D=\lim_{\epsilon \to 0} \frac{\log N(\epsilon)}{- \log \epsilon}
\end{equation}
by least squares fitting. We display an example of application of box counting in Fig. \ref{fig_FD}.

For the case of grayscale images or continuous functions, we generated 8 binary sets using multiple Otsu thresholding, then computed the fractal dimension, area, and mean intensity for each point set as in \cite{Costa2012}.

\paragraph{Wavelet Texture Descriptors}   
A multi-scale texture descriptor is usually very useful for classification. Gabor filter banks and wavelet transforms are both multi-scale spatial-spatial frequency filtering techniques. The discrete wavelet transform is frequently applied  using tree or pyramid hierarchies for texture representation. Multi-band analysis offers advantages over the traditional discrete Fourier transform, but wavelet transform does not produce as exact a result as the Fourier transform.

\paragraph{Discrete Wavelet Frames}
Discrete wavelet frames employ a filter bank for multi-scale decomposition. The Haar wavelet with a low-pass filter 
\begin{equation}
H(z) = (1+z^{-1})/2
\end{equation}
and a corresponding high-pass filter
\begin{equation}
G(z) = (1-z^{-1})/2
\end{equation}
is frequently used because of its efficiency and computational simplicity. 

The largest filter kernels will have size $2^{maxlevel}$, where the $maxlevel$ is the number of multiresolution levels. At each level, we filter the image by using the filter combinations:
\begin{equation}
H_x H_y,\,\, H_x G_y,\,\, G_x H_y,\,\, G_x G_y,
\end{equation} 
where $H_x$ is the low-pass filter along the $x$ direction, and $G_y$ is the high-pass filter along the $y$ direction. 

To produce the wavelet frame representation we compute the discrete wavelet transform for all possible signal shifts at multiple scales. The filters are used to decompose the image into subbands. We compute the orthogonal projections and residuals for a full discrete wavelet expansion. We then compute energy, variance, entropy, contrast, skewness, and kurtosis signatures to form the texture descriptor. These characteristics are calculated as follows.
\paragraph{Contrast}
It measures the intensity contrast between a pixel $p(i,j)$ and its neighbors in an image by
\begin{align}
\sum_{i,j}{|i-j|^2 p(i,j)}
\end{align}
\paragraph{Energy}
It is expressed by the sum of squared elements 
\begin{align}
\sum_{i,j}{p(i,j)^2} 
\end{align}
\paragraph{Skewness}
It measures the lack of symmetry. For a random variable $x$, the skewness is the third standardized moment $\gamma_1$ 
\begin{align}
\gamma_1 = \text{E}\left[ \left( \frac{X - \mu}{\sigma} \right)^3 \right] = \frac{\mu_3}{\sigma_3} = \frac{\text{E}[(X-\mu)^3]}{(E[(X-\mu)^2])^{3/2}} = \frac{\kappa_3}{\kappa_2^{3/2}}
\end{align}
where $\mu$ is mean, $\sigma$ is standard deviation, $\mu_3$ is central moment, E is expectation operator and $\kappa_i$ is the $i$th cumulants.
\paragraph{Kurtosis}
It measures the degree to which data points follow a heavy-tailed or light-tailed distribution. Higher kurtosis values, correspond to heavier-tailed distributions.
\begin{align}
Kurt[X] = E\left[ \left( \frac{X-\mu}{\sigma} \right)^4 \right] = \frac{mu_4}{\sigma^4} = \frac{\text{E}[(X-\mu)^4]}{(\text{E}[(X-\mu)^2])^2}
\end{align}

\paragraph{Entropy}
It presents the state of a system, such as the disorder and randomness of the system. The wavelet entropy is defined in \cite{Blanco1998} as
\begin{equation}
S(p) = - \sum_{j}{p_j \dot \ln{p_j}}
\end{equation}

\begin{figure}[H]
\centering
\includegraphics[width=.65\columnwidth,clip,trim= 1.1in 0.9in 0.8in 0.6in]{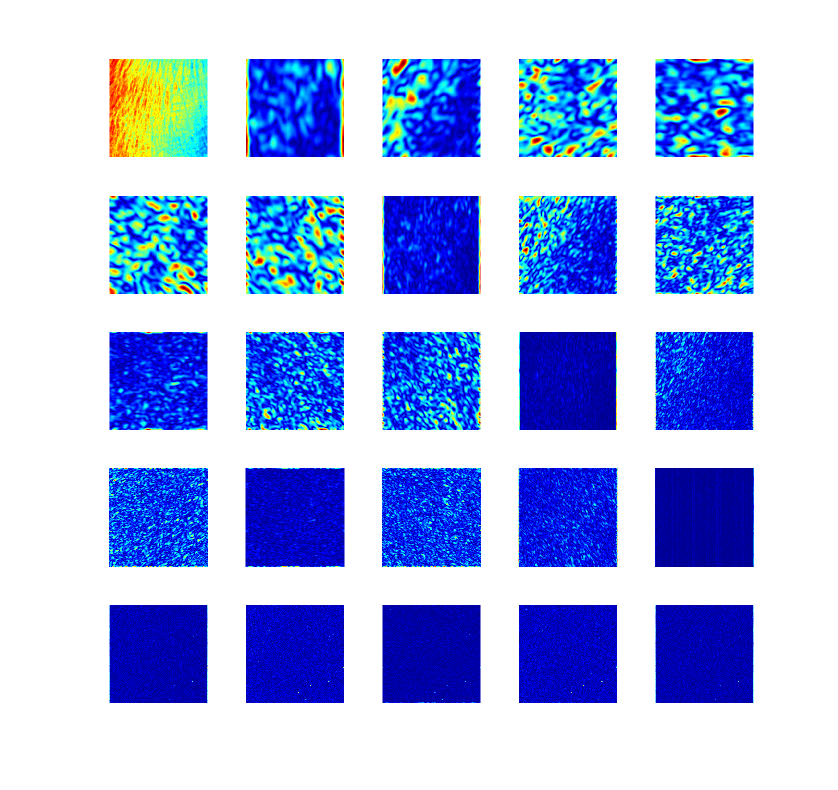}
\caption{The original bone radiograph and the Gabor texture components of a healthy subject using 4 scales and 6 orientations. The 24 components are calculated on the original image in top left. While these maps pronounce the texture characteristics, visual interpretation is still particularly challenging. Therefore a machine learning technique is needed to distinguish healthy from osteoporotic subjects.}
\label{fig_Gabor}
\end{figure}

\paragraph{Wavelet Gabor Filter Bank}
The Gabor filter is a linear filter that can extract relevant characteristics for multiple frequencies and orientations (Fig. \ref{fig_Gabor}), similarly to the human visual system. 

Gabor functions form a complete but non-orthogonal basis. In the spatial domain, a 2D Gabor filter is a Gaussian kernel function modulated by a sinusoidal plane wave. Gabor filters are often used for texture identification, and good results have been achieved. The filter is represented in complex form as follows:
\begin{equation}
g(x,y;\lambda,\theta,\psi,\sigma,\gamma) = \exp \left(-\frac{w^2 + \gamma^2 v^2}{2 \sigma^2} \right) \exp \left( i (2 \pi \frac{w}{\lambda} + \psi) \right)
\end{equation}
and
\begin{align}
w &= x \cos \theta + y \sin \theta\\
v &= - x \sin \theta + y \cos \theta
\end{align}
where $\lambda$ is the wavelength of the sinusoidal factor, $\theta$ is the orientation of the normal to the parallel stripes of a Gabor function, $\psi$ is the phase offset, $\sigma$ is the standard deviation of the Gaussian envelope and $\gamma$ is the spatial aspect ratio. 
 
The filter dictionary can be produced by dilations and rotations of the mother Gabor wavelet.

\paragraph{Local Binary Patterns (LBP)}    
For each pixel $pix$ in the image, we compare the intensity of $pix$ to the intensities of its eight neighbors. If the intensity of $pix$ is greater or equal to its $i$th (where $i = 1,2,...,8$) neighbor, we set $b_i = 0$, otherwise $b_i = 1$. From these eight neighbors we construct an eight-digit binary number $b_1b_2b_3b_4b_5b_6b_7b_8$. We use the histogram of these numbers as a texture descriptor \cite{Shapiro2001}.

\paragraph{Discrete Fourier and Cosine Transforms}   
The Discrete Fourier transform and the Discrete Cosine transform coefficients aim to capture characteristics of texture in the spatial frequency domain. For example, fine texture has greater high frequency components, whereas coarse texture is represented by lower frequencies. The Discrete Fourier and Cosine transforms are defined as follows,

Discrete Fourier transform (DFT): 
\begin{equation}
F(k,l) = \frac{1}{MN}\sum_{m=0}^{M-1}\sum_{n=0}^{N-1}{f(m,n) \cdot e^{-j 2 \pi (\frac{mk}{M} + \frac{nl}{N})}},
\end{equation}
where $f(m,n)$ is the pixel intensity at $(m,n)$, and $k = 0,1,2,...,N-1,l = 0,1,2,...,M-1$.

Discrete Cosine Transform (DCT) uses only cosine basis functions:
\begin{equation}
C(k,l) = \sqrt{\frac{\alpha}{MN}} \sum_{m=0}^{M-1} \sum_{n=0}^{N-1} {f(m,n) \cdot \cos{\frac{\pi (2m + 1) k}{2M}} \cos{\frac{\pi (2n + 1) l}{2N}}},
\end{equation}
where $\alpha=1, \text{if } k=l=0; \alpha = 4, \text{if } 1 \leq k \leq M-1, 1 \leq l \leq N-1$.

We use the $8 \times 8$ coefficients corresponding to lower frequencies for classification.
	
\paragraph{Law's Texture Energy Masks}  
The texture energy is computed by a set of $5 \times 5$ convolution masks (level, edges, waves, spots, and ripples) to measure the amount of variation within a fixed-size window. We use the average level (intensity) feature to normalize intensity range and then we use the remaining 24 components to form the texture vector. Next, we calculate the mean, variance, energy, skewness, kurtosis, and entropy for each component.
	
\paragraph{Edge Histogram}    
We compute the intensity gradient magnitude $\vert \nabla f \vert$ and then calculate its histogram by 
\begin{align}
p_{|\nabla f|} \left(|\nabla f| = r_k \right) &= \frac{n_k}{N},\,\,\,k = 0,...,L-1 \\
\nabla f &= \left( \frac{\partial f}{\partial x_1}, \frac{\partial f}{\partial x_2}, \cdots, \frac{\partial f}{\partial x_N} \right)^\top
\end{align}

\begin{figure}[t]
\centering
\includegraphics[width=1\textwidth,clip,trim= 0in 0in 0in 0in]{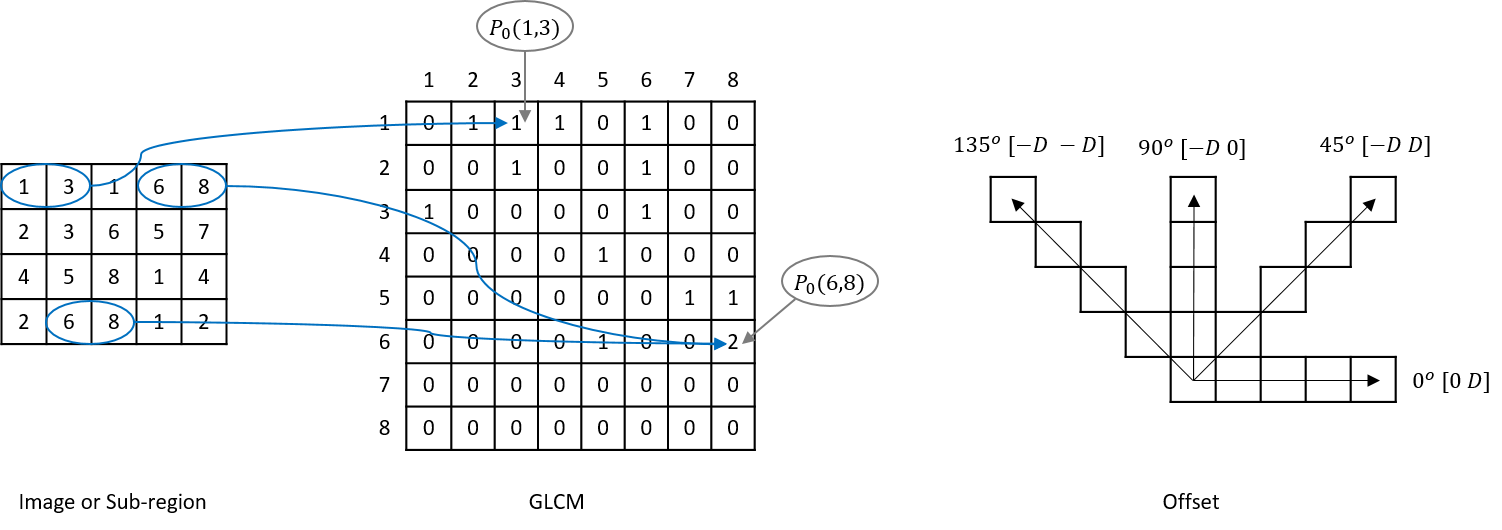}
\caption{Process used to create the GLCM (left) and Offset of GLCM (right)}
\label{fig_GLCM_Offset}
\end{figure}

\paragraph{Gray Level Co-Occurrence Matrix (GLCM)}    
The GLCM calculates how the frequency of occurence of gray-level pairs $(i,j)$ in horizontal, vertical, or diagonal pixel adjacencies on the image plane, displayed in Fig. \ref{fig_GLCM_Offset}. Horizontal ($0^\circ$), vertical ($90^\circ$), and diagonal ($-45^\circ, -135^\circ$) dimensions of analysis are denoted by $P_0, P_{90}, P_{45},$ and $P_{135}$, respectively. After we create the GLCMs, we compute contrast, correlation, energy and homogeneity measures.

\subsubsection{Feature Selection}
\label{sec_feature_selection}
This classification component aims to select relevant and informative features for classification. It is applied to improve classification performance, to reduce computational complexity, and to interpret data.

\paragraph{Correlation-based Feature Selection (CFS)}   
This method selects features that are highly correlated with the pattern classes, but have low correlation with the remaining features. The subset evaluation function is given by:
\begin{equation}
\label{eq_cfs}
Merit_S = \dfrac{\bar{k}_{r_{cf}}}{\sqrt{k+k(k-1) \bar{r_{ff}}}}
\end{equation}
where $Merit_S$ is the merit of the selected feature set $S$, $\bar{k}_{r_{cf}}$ is the mean correlation between the features and class with $f \in S$, and $\bar{r_{ff}}$ is the mean pairwise feature correlation. The numerator expresses predictive capacity, while the denominator expresses feature redundancy.

\paragraph{Best First Search (BF)}    
Searches the space of feature subsets by greedy hillclimbing that may include backtracking. Best first may search forward, or backward, or, consider all possible single feature additions and deletions at a given point using a bi-directional strategy.

\paragraph{Genetic Algorithm-based Search (GA)}    
Genetic search works by having a population of variables representing feature sets and performs the operations of reproduction, cross-over and mutation in each generation to get the offspring that optimizes a feature set-related objective function.

\paragraph{Information Gain (IG)}    
This function measures the information gain with respect to the class:
\begin{equation}
{\footnotesize \text{InfoGain(Class,Attribute)} = H(\text{Class}) - H(\text{Class} \vert \text{Attribute)}}
\end{equation}
where $H$ is the entropy of each class given by $H(\text{Class}) = - p_{\text{Class}} \log p_{\text{Class}}$
We select the attributes by individual ranking evaluation.

\paragraph{Ranker}
Using Ranker as a search means that we will rank the features based on the features' individual evaluations. A threshold can be set in Ranker, and features that are smaller than this threshold will be removed from the feature set. Ranker is used with attribute evaluators, such as Information Gain (IG), feature selection and entropy, etc. 

\subsubsection{Classifiers and Discriminant Functions}
\label{sec_texture_classifiers}
\paragraph{Na\"{i}ve Bayes (NB)}   
This model assumes conditional statistical independence
	$p(x|\omega_j) = \prod_{k=1}^{D}{p(x_k|\omega_j)}$
where $x = (x_1, x_2, \ldots, x_D)^T$ and $D$ is the dimensionality of the feature space. The posterior probability is based on Bayes' formula:
	$$p(\omega_j|x) = \frac{p(x|\omega_j) P(\omega_j)}{p(x)}.$$
 The MAP decision rule is typically used for classification.
Suppose we have two categories $\omega_1$ and $\omega_2$ with discriminant functions $g_1(x),g_2(x)$, where
	$$g_i(x)=-\frac{1}{2}(x-\mu_i)^T\Sigma_i^{-1}(x-\mu_i) + \frac{D}{2}\ln{2\pi}-\frac{1}{2}\ln{|\Sigma_i|} + \ln{P(\omega_i)}.$$
Then we can define a single discriminant function by
	$$g(x) = g_1(x) - g_2(x).$$
The decision rule is :
\[
\begin{cases}
    \omega_1,& \text{if} \,\, g(x)>0\\
    \omega_2,& \text{if} \,\, g(x)<0
\end{cases}.
\]
\paragraph{Multilayer Perceptron (MLP)}    
A multilayer perceptron is a feedforward artificial neural network system that maps input patterns onto class labels. An MLP has multiple layers of nodes that are fully connected to the next layer. Each node is a neuron with a nonlinear activation function. MLP utilizes backpropagation for supervised learning \cite{rumelhart1986learning,Duda2001}. Because MLP has multiple layers of logistic regression models, it can distinguish data that are not linearly separable. In learning by backpropagation -that can be considered as an extension of the LMS algorithm- we adjust the connection weights, according to the amount of error in the output compared to the expected result.

\paragraph{Bayes Network (BN)}    
A Bayes network, is a probabilistic graphical model that uses  a directed acyclic graph to represent a set of random variables and their conditional dependencies. In a Bayesian network the joint probability density function can be written as the product of univariate conditional density functions dependent on their parent variables:
\begin{align}
p(x) = \Pi_{v \in V}{p\left(x_v|x_{pa(v)}\right)}
\end{align}
where $pa(v)$ denotes the parents of $v$. In the graph, the parents are vertices directly connected to $v$ by a single edge.
		
\paragraph{Bagging}    
For a training set $S$ with size $k$, Bagging generates $j$ training subsets denoted as $S_i$ with size $k'<k$, by sampling from $S$ uniformly and with replacement. We denote the original set as $A$. In the training stage, we first have $D=\emptyset$ and $j$ is the number of classifiers to train. Then for $p = 1,2,...,j$, we take a bootstrap sample $S_p$ from $A$ to train classifier $D_p$. Then the classifier $D_p$ is added to the current ensemble, $D=D \cup D_p$. The class label prediction for the input $x$ is obtained by majority voting on the individual classifier decisions produced by $D_1,...,D_j$ \cite{Duda2001}.

\paragraph{Random Forests (RF)}    
Random forests is an ensemble learning method that constructs multiple decision trees from subsets of the training set and uses random feature selection for node splitting. RF decides the class after applying voting to the predicted classes by the individual trees for classification, or by calculating the mean prediction for regression. Random forests address the overfitting tendency of the decision trees and have shown robustness with respect to noise \cite{Mitchell1997}.

\subsection{Bag of Keypoints}
Bag of Keypoints (BoK) \cite{Csurka2004} is a patch-based technique that originates from Bag of Features methods. These methods have been applied to image recognition and classification and have produced very good results. They apply feature detection, extraction and clustering for finding the most representative features in the training database. In the next step, they build a vocabulary that consists of the frequency of occurrence of these features. In the testing stage, features are extracted from the unlabeled image and encoded using the vocabulary that was built during training. Then a learning method is applied to classify the test pattern into one of the classes. 

We employed the support vector machine (SVM) classifier for learning a discriminant function from the encoded features and classifying unlabeled samples. In the SVM module we evaluated the use of linear or radial basis function kernels. We utilized radial basis function kernels for our experiments to address possible non-linearity of the decision boundary. 

\subsection{Deep Neural Networks}
\label{sec_DeepLMethods}
Deep learning methods and more specifically convolutional neural networks have recently re- emerged as powerful techniques for image segmentation, object recognition and classification \cite{Krizhevsky2012imagenet,Szegedy2015,Szegedy2016,Tajbakhsh2016,He2016}. These techniques simultaneously learn the set of features and the decision function. In contrast to traditional texture-based techniques, deep networks do not need to receive a hand-crafted feature set as input. Deep learning methods have been applied to biomedical image datasets and have produced very good results.

We employed sequential and residual networks of varying complexity such as Alexnet \cite{Krizhevsky2012imagenet}, Googlenet \cite{Szegedy2015}, Resnet18 \cite{He2016}, and Inceptionv3 \cite{Szegedy2016}. Because our datasets are small, we employed transfer learning techniques to adjust the weights of pre-trained networks, instead of learning the decision function from the beginning as described in \cite{Tajbakhsh2016,Shin2016}. All networks were pre-trained on Imagenet that is a database of 1.2 million natural images.

We applied transfer learning to each network in slightly different ways.
To adjust Alexnet to our data, we replaced the pre-trained fully connected layers with three new fully connected layers. We set the learning rates of the pre-trained layers to 0 to keep the network weights fixed, and we trained the new fully connected layers only. In the case of Googlenet, we set the learning rates of the bottom 10 layers to 0, we replaced the top fully connected layer with a new fully connected layer, and we assigned a greater learning rate factor for the new layer than the pre-trained layers.

To provide the networks with additional training examples, we employed data resampling using randomly-centered patches, followed by data augmentation by rotation, scaling, horizontal flipping, and vertical flipping.
Finally, we applied hyperparameter tuning using Bayesian optimization to find the optimal learning rate, mini-batch size and number of epochs.

\subsection{Sparse Representation and Classification}
\label{ch_sparsity_base_technique}
The concept of sparsity has been used in many methods of mathematics, computer science and engineering and plays an important role in machine learning and pattern recognition. Next, we introduce the standard sparse technique, the details of this method, and other related sparse techniques.
 
\subsubsection{Overview of Sparse Modeling Methods}
\label{sec_Overview_of_Sparse_Modeling_Methods}
Tissue classification is typically achieved by supervised machine learning approaches. Among numerous techniques that proposed generative or discriminative models, use of kernels, and linear or nonlinear approaches, sparse classification techniques have shown promise and applicability for characterizing visual patterns in region of interest (ROI)-based analyses. Sparse representation techniques have been applied to extensive fields including coding, feature extraction and classification, superresolution \cite{Yang2008}, and regularization of inverse problems
\cite{Figueiredo2007}. Exploration of signal's sparsity may provide insight into the important patterns of prototyping of objects category.
The sparse representation is more concise for compression and naturally discriminative for classification \cite{Wright2009}. 
Sparse representation techniques calculate a sparse linear combination of atoms for describing a vector sample using an overcomplete dictionary of prototypes.
If the representations of these linear combinations are sufficiently sparse, then they can be used for object recognition and classification of imaging patterns.

The authors in \cite{Wright2009} proposed the sparse representation classification (SRC) method to recognize 2,414 frontal-face images of 38 individuals of Yale B Database and over 4,000 frontal images for 126 individuals of AR Database, producing recognition rates greater than 90\% for both databases. Another notable face recognition application of sparse coding was published in \cite{Qiao2010} reporting high levels of classification accuracy. 
Dictionary learning techniques have also emerged as solutions for sparse representation in the recent years. The utilization of K-SVD, where SVD denotes singular-value decomposition, for dictionary learning has been studied to produce a dictionary aiming for more accurate representation \cite{Aharon2006}. In \cite{Mairal2008}, the K-SVD technique has been used for color image restoration to handle nonhomogeneous noise and information missing problems. The authors in \cite{Zepeda2009} observed that K-means may yield as good precision rate as K-SVD when we use the same number of atoms. 
The SRC method with dictionary learning was applied to classification of pulmonary patterns of diffuse lung disease in \cite{Zhao2015}. Additional algorithms such as matching pursuit (MP), orthogonal matching pursuit (OMP), and basis pursuit (BP) have been proposed for codebook design \cite{Aharon2006}.

\subsubsection{Sparse Representation and Classification}
\label{sec_Sparse_Representation_and_Classification}
Sparse representation or approximation techniques construct a dictionary from labeled training samples to calculate a linear representation of a test sample. This representation can be used to make a decision for the class of the test sample. Assuming that a dataset has $k$ distinct classes, $s$ samples, and for $i$th class there are $s_i$ samples, so that $s = \sum_{i}{s_i}$, we define a dictionary matrix $M$ from the training set as
\begin{equation}
	M = [v_{1,1}, v_{1,2}, ..., v_{k,s_k}].
\end{equation}
where $M \in \mathbb{R}^{l \times s}$, and $v_{i,h}$ is a column vector for the $h$th sample from $i$th class. In image classification applications, a $p \times q$ grayscale image forms a vector $v \in \mathbb{R}^{l}$, $l = p \times q$ using lexicographical ordering.

A new test sample $y \in \mathbb{R}^{l}$, can be represented by a linear combination of samples $y = \sum_{i=1}^{k}{\beta_{i,1}v_{i,1} + \beta_{i,2}v_{i,2} + \cdots + \beta_{i,s_i} v_{i,s_i}}$, where $\beta_{i,h} \in \mathbb{R}$ are scalar coefficients.
Hence, the test sample $y$ can be rewritten as:
\begin{equation}
	y = Mx_0 \in \mathbb{R}^l.
\end{equation}
where $x_0$ is a sparse solution. If there are sufficient training samples, the components of $x_0$ are equal to zero except for the components corresponding to the $i$th class. Then $x_0 = [0,0,...,\beta_{i,1},\beta_{i,2},...,\beta_{i,s_i},0,0,...,0]^T \in \mathbb{R}^s$.

In \cite{Donoho2003}, it was proved that whenever $y=Mx$ for some $x$, if there are less than $l/2$ nonzero entries in $x$, $x$ is the unique sparse solution: $\widehat{x}_0 = x$.
Finding an accurate sparse representation of an underdetermined system of linear equations is an NP-hard problem \cite{Davis1997}, therefore only approximate solutions can be found. The authors in \cite{Donoho2004} supported that if the solution $x_0$ is sparse enough, it is equal to the solution $\widehat{x}_1$ of the $l^1$-minimization problem:
\begin{equation}
	(l^1):\quad \widehat{x}_1 = \arg\min ||x||_1\quad s.t. \quad Mx = y.
	\label{eq_l1}
\end{equation}

In sparse representation classification we define a characteristic function $\delta_i : \mathbb{R}^s \rightarrow \mathbb{R}^s$  that has nonzero entries, only if $x$ is associated with class $i$. Then the function $\hat{y}_i = M \delta_i(\hat{x}_1)$, represents the given sample $y$ using components from class $i$ only. To classify $y$ and determine the class label $\widehat{\omega_i}$, we minimize the residual between $y$ and $\hat{y}_i$ \cite{Wright2009}: 
\begin{equation}
	\widehat{\omega_i} = \arg \min_{i} {r_i(y)} \doteq ||y - M\delta_i(\hat{x}_1)||_2.
\label{eq_SRC_residual}
\end{equation}
This technique also adopts the sparsity concentration index (SCI) to measure the efficiency of class-conditional representation of a sample. The SCI of a coefficient vector $x \in \mathbb{R}^s$ is $SCI(x) = \frac{k \times \max_i ||\delta_i(x)||_1/||x||_1 - 1}{k - 1} \in [0,1]$ as defined in \cite{Wright2009}. For a solution $\widehat{x}$, if $SCI(\widehat{x})$ is 1, $y$ is only represented by images from a single class, and if $SCI(\widehat{x}) = 0$, the components of $\beta $ are spread evenly over all classes.

\subsubsection{Algorithms for Solving the Sparse Representation Problem}
\label{sec_Sparse_NP_Hard_Problem}
Earlier in this section we mentioned that finding an accurate solution of sparse representation is an NP hard problem, and described a method for finding an approximate solution to Eq. \eqref{eq_l1}. Here we outline three common methods for solving \eqref{eq_l1}. One is the matching pursuit (MP) method. MP selects atoms, one at a time, to minimize the approximation error. This is done by finding the atom with the largest inner product of the signal, subtracting the approximation from the signal using only that atom, repeat this step until it finds the satisfying residual. In \cite{Pati1993} the authors proposed a algorithm as orthogonal matching pursuit (OMP). OMP modified MP to achieve full backward orthogonality of residuals (errors) at each step, resulting in improved convergence. Another optimization method for this problem is basis pursuit (BP) \cite{Aharon2008}. This method optimizes a joint expression of the contstraint and the objective function and is equivalent to the LASSO method. 

\subsubsection{Second Order Cone Programming Formulation}
The second order cone (SOCP) programming problems are convex optimization problems. The SOCP can be used to implement linear programming (LP), convex quadratic programs (QPs) and convex quadratically constrained quadratic programs (QCQPs) \cite{Alizadeh2001}. The standard form of SOCP is defined as following:
\begin{align}
\min \,\,\,\,\,\,\,\, & u_1^{\top}x_1 + \cdot\cdot\cdot + u_n^{\top}x_n \\
s.t. \,\,\,\,\,\,\,\, & A_1x_1 + \cdot\cdot\cdot + A_n x_n = b\\
					  & x_i \geq 0 \text{ for } i = 1,2,...,n
\end{align}
SOCP can be used for solving problems of the form,
\begin{align}
\min_x{f(x)} \text{ s.t. } 
  \begin{cases}
    c(x) \leq 0 \\ 
    ceq(x) = 0  \\
    A \cdot x \leq b\\
    Aeq \cdot x = beq\\
    lb \leq x \leq ub
  \end{cases}
\label{eqt_SOCP_fmincon}
\end{align}
where $f(x)$ is an objective function, $lb$ and $ub$ are lower bound and upper bound respectively, $A$ is a matrix and $b$ is a vector for inequality, $Aeq$ is a matrix and $beq$ is a vector for equality, and $c(x)$ and $ceq(x)$ are constraint functions that return vectors. Especially, $f(x), c(x)$ and $ceq(x)$ can be nonlinear functions.\\

\subsection{Integrative Ensemble Sparse Analysis Techniques}
\begin{figure}[tp]
\centering
\includegraphics[width=1 \textwidth,clip,trim=0in 0in 0in 0in]{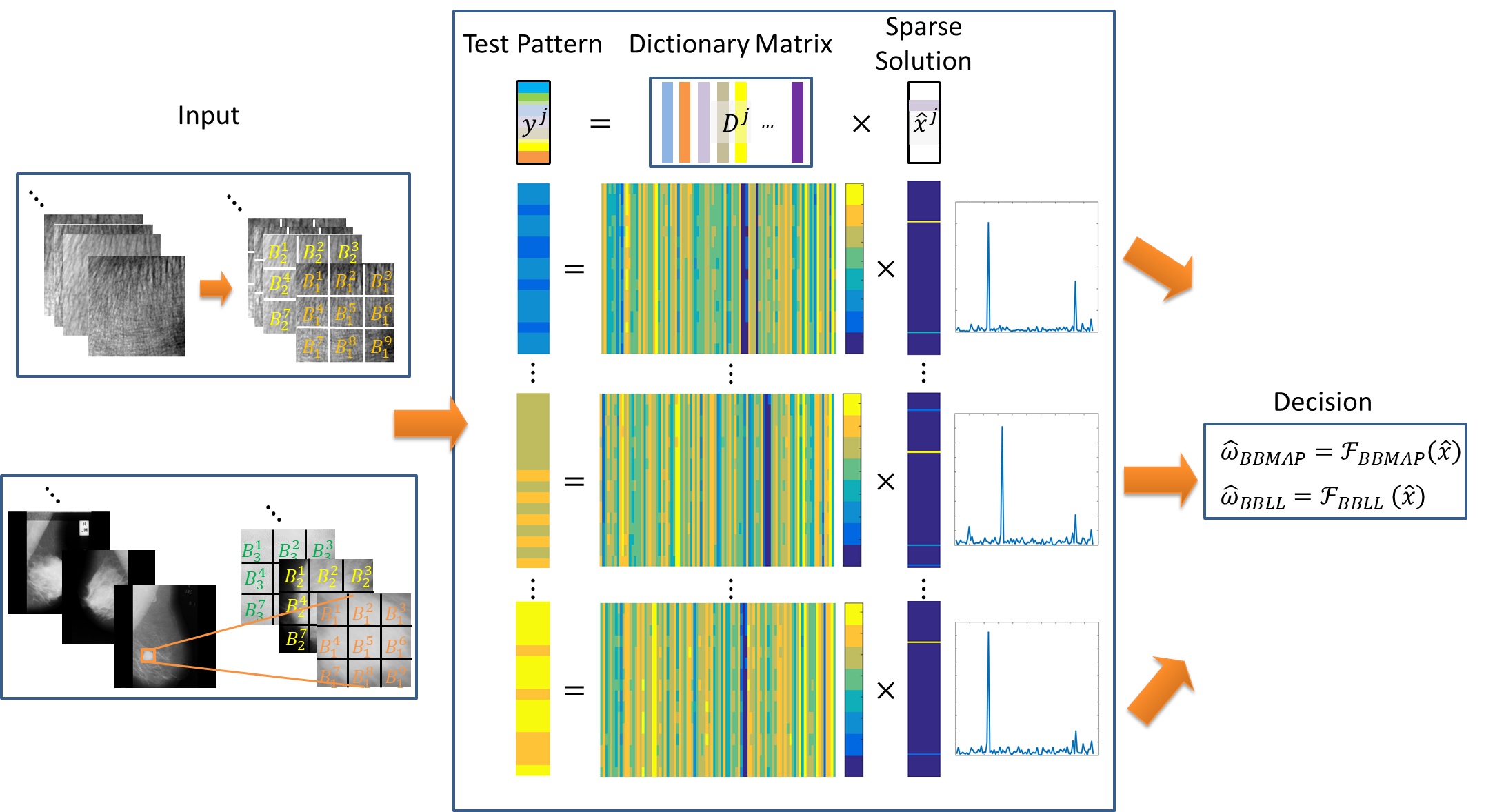}
\caption{Main stages of our integrative sparse modeling system: block-based analysis, sparse solutions, and decision functions.}
\label{fig_Graphical_Abstract}
\end{figure}

This method builds an ensemble of sparse representation classifiers based on block decomposition of the input ROI to address shortcomings caused by high dimensionality and to introduce spatial localization in sparse approximations \cite{Zheng2020}. Fig. \ref{fig_Graphical_Abstract} summarizes the main stages of our method that may be divided in block-based learning and Bayesian model averaging to form decision functions.

\subsubsection{Block Decomposition}
\label{sec_intoBB}
We first divide each training ROI into non overlapping blocks of size $ m \times n $.
Thus, each ROI image is expressed as $I = [B^1, B^2, ... ,B^{NB}]$, where $NB$ is the number of blocks in an image. 
The dictionary $ D^j $, where $ j = 1,2, ..., NB $ corresponds to the block $B^j$ at the same index within the image ROI.
The dictionary $ D^j $ for all the $s$ images can be represented as follows:

\begin{equation}
	D^j = [{bv}_{1,1}^j, {bv}_{1,2}^j, \ldots , {bv}_{k,s_k}^j],
\end{equation}
where ${bv}_{i,h}^j$ is the column vector denoting the $h$th sample, $i$th class, $j$th block $B^j$.

\subsubsection{Ensemble Classification}
\label{sec_BB}
In this stage, each test sample is classified by constructing ensembles of classifiers that solve a set of sparse coding and classification problems, or hypotheses corresponding to the block components.
Given a test sample $y^j$ in $j$th block, we find the solution $x^j$ of the regularized noisy $l^1$-minimization problem:
\begin{equation}
	\widehat{x}^j = \arg \min {||x^j||_1}\,\, \mbox{subject to}\,\, \Vert D^j x - y^j \Vert_2 \leq \epsilon 
	\label{eqt_block_src}
\end{equation}
where $j = 1,2,...,NB$. 
The test sample $y^j$ will be assigned to the class $\omega^j_i$, which has minimum approximation error calculated by \eqref{eqt_l1_deci}.
\begin{equation}
\omega_i = \arg \min_{i} {r_i(\widehat{x})} \doteq \arg \min_{i} \Vert y - \widehat{y}_i \Vert_2.
\label{eqt_l1_deci}
\end{equation}

We utilize ensemble learning techniques in a Bayesian probabilistic setting as weighted sums of classifier predictions. We employ a function that applies majority voting to individual hypotheses (BBMAP) and an ensemble of log likelihood scores computed from relative sparsity scores (BBLL). 
\label{sec_BBdeci}

\paragraph{Maximum a Posteriori decision function (BBMAP)}
The class label for each test sample is determined by voting over the ensemble of $NB$ block-based classifiers. The predicted class label $\widehat{\omega}$ is given by
\begin{equation}
	\widehat{\omega}_{BBMAP} = \mathcal{F}_{BBMAP}(\widehat{x}) \doteq \arg{\max_i{pr(\omega_i|\widehat{x})}},
\label{eqt_BBMAP}
\end{equation} 
where $\widehat{x}$ is the composite extracted feature from the test sample given by the solution of \eqref{eqt_block_src}. The probability for classifying $\widehat{x}$ into class  $\omega_i$ is 
\begin{align}
pr(\omega_i|\widehat{x}) &= \sum_j^{NB} ND_{\omega^j_i} / NB \\
ND_{\omega^j_i} &=
\begin{cases}
 1, & \text{if } \widehat{x}^j \in i\text{th class} \\
 0, & \text{otherwise}
			   \end{cases},
\end{align}
where $ND_{\omega^j_i}$ is an indicator function whose values are determined by the individual classifier decisions.

\paragraph{Log likelihood approximation residual-based decision function (BBLL-R)}
We define a likelihood score based on the residuals $r_m,r_n$ calculated in the sparse representation stage of each classifier. We calculate the expectation of $\overline{LLS(\widehat{x})}$ over all individual classifiers:
\begin{equation}
\overline{LLS(\widehat{x})} = -\frac{1}{NB}  \left[ \sum_j^{NB} \log{r_m^j(\widehat{x})} - \sum_j^{NB} \log{r_n^j(\widehat{x})} \right],
\end{equation} 
where $r_\omega^j(y)$ is the approximation residual for class $\omega$ and $j$ is the block index:
\begin{equation}
r_\omega^j(y) = ||y - M \delta_\omega(\hat{x}_1)||_2 \text{ for } j = 1, ..., k. 
\label{eqt_BBLL_residuals}
\end{equation}

\paragraph{Log likelihood sparsity-based decision function (BBLL-S)}
We define a likelihood score based on the relative sparsity scores $\Vert \delta_m ( \widehat{x}^j ) \Vert_1$, $\Vert \delta_n ( \widehat{x}^j ) \Vert_1$ calculated in the sparse representation stage of each classifier. We calculate the expectation of $LLS(\widehat{x})$ over all individual classifiers:
\begin{equation}
\overline{LLS(\widehat{x})} = -\frac{1}{NB}  \left[ \sum_j^{NB} \log{\Vert \delta_m ( \widehat{x}^j ) \Vert_1} - \sum_j^{NB} \log{\Vert \delta_n ( \widehat{x}^j ) \Vert_1} \right],
\end{equation} 
The introduction of the log-likelihood score accommodates the definition of a decision function for the state $\widehat{\omega}$. To determine the class we apply a decision threshold $\tau_{LLS}$ to $\overline{LLS(\widehat{x})}$. 
\begin{equation}
\widehat{\omega}_{BBLL} = \mathcal{F}_{BBLL}(\widehat{x}) \doteq \begin{cases} 
					m\text{th class}, & \text{if } \overline{LLS(\widehat{x})} \geq \tau_{LLS} \\
     				n\text{th class}, & \text{otherwise } 
    		  	   \end{cases}.
\end{equation}

\begin{figure}[thpb]
\centering
\includegraphics[width=0.48\textwidth,clip,trim=.15in 0in .5in .1in]{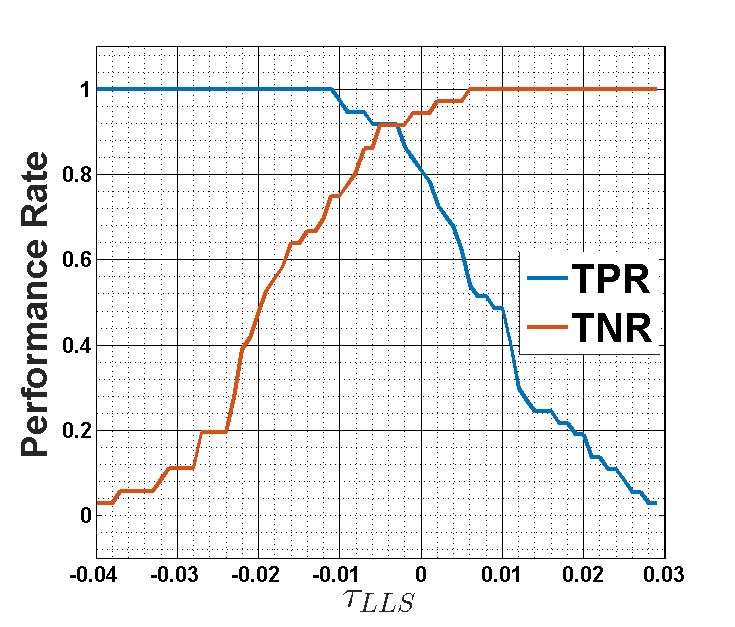}
\includegraphics[width=0.48\textwidth,clip,trim=.15in 0in .5in .1in]{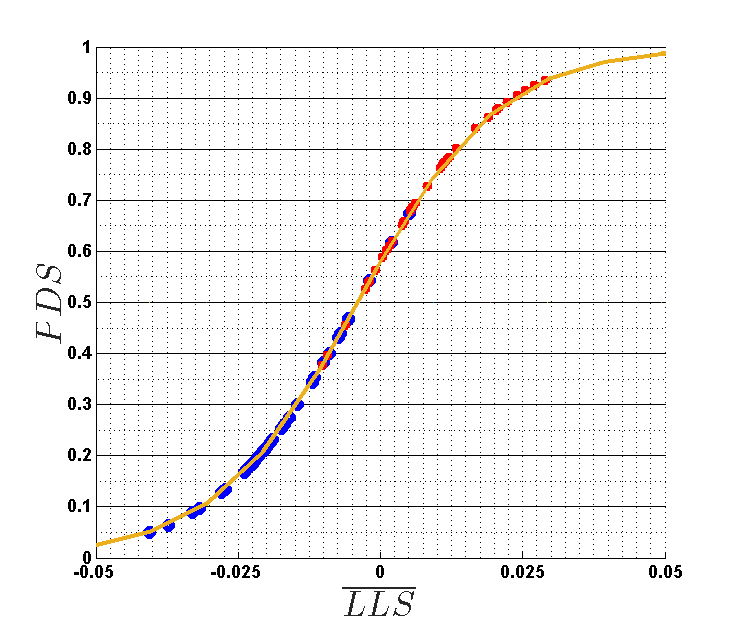}
\caption{\label{fig_MIAS_BB_SRC_LP_threshold} An example of TPR and TNR curves versus $\tau_{LLS}$  for determining $\tau^*_{LLS}=c$ (left) and the sigmoid probability decision score $PDS$ after calculating the parameters $m,c$ for \eqref{eqt_PDS} (right).}
\end{figure}

This threshold is expected to be equal to 0, if there is no estimation bias, but may be experimentally determined as the minimizer of a Bayes-type risk function. Hence the optimal $\tau^*_{LLS}$ value can be determined by sampling the domain of $\tau_{LLS}$ and calculating true positive and true negative rates. Next, the optimal value is determined by the intersection of TPR and TNR curves. An example of this procedure for determining $\tau^*_{LLS}$ is displayed in Figure \ref{fig_MIAS_BB_SRC_LP_threshold}.

In the next stage, we aim to convert the log likelihood decision scores to bounded posterior probability values using a sigmoid function. This function is denoted by Probability Decision Score ($PDS$) and is expressed by
\begin{equation}
\label{eqt_PDS}
PDS(\overline{LLS}) = \frac{1}{1+\exp\left[-m(\overline{LLS}-c)\right]}
\end{equation}
To calculate the model parameter $c$, we require that this function be equal to 50\% probability for $\tau^*_{LLS}$, hence $c=\tau^*_{LLS}$. To estimate $m$, we set a fixed probability level $PDS_{min}$ (e.g., 5\%, 10\%) for the smallest value $\overline{LLS}_{min}
$.

\begin{equation}
m = \frac{1}{\tau^*_{LLS} - \overline{LLS}_{min}} \ln \left( \frac{100 - PDS_{min}}{PDS_{min}} \right)
\end{equation}

In Figure \ref{fig_MIAS_BB_SRC_LP_threshold} we display the graph of $PDS$ versus $LLS$ for one experiment. We can use $PDS$ to express margins of uncertainty for classification in percentiles.


\section{Results}
The main goal of the presented experiments is to test the hypothesis that ensembles of block-based sparse classifiers improves the classification performance of conventional sparse representation. The second goal is to compare the proposed technique to classifiers based on texture, Bag of Keypoints, and deep learning. Finally, we compare the performances of the two decision functions BBMAP and BBLL. We describe the application of our system to osteoporosis diagnosis, we report the classification results produced by leave-one-out cross-validation experiments, and discuss our findings.

\paragraph{Data Description}
The objective is to distinguish between healthy and osteoporotic subjects. The TCB challenge dataset contains labeled digital radiographs of 87 healthy and 87 osteoporotic subjects for training and testing (available online in  {http://www.univ-orleans.fr/i3mto/data}, last access in 05/2018). The calcaneus trabecular bone images in the dataset have an ROI size of $400 \times 400$ pixels. A more detailed description of the dataset is provided in \cite{Hassouni2017}. The experimental procedures involving human subjects were approved by the Institutional Review Board of the institution that provides the data. 

\paragraph{Texture-based Classification}
In the performance evaluation of conventional texture-based techniques, we calculated 723 texture-related features \cite{Zheng2016}. We selected features using correlation-based feature selection with best first search (CFS-BF), correlation-based feature selection with genetic algorithm search (CFS-GA), information gain (IG) and no-feature selection as described in Section \ref{sec_feature_selection}.
CFS-GA yielded an overall better performance than CFS-BF, IG and no-feature selection on leave-one-out cross-validation (Table \ref{tab_TCB_BB_Weka_MatLab_BB_LOO}). This implies that CFS-GA effectively selects distinguishing features from the entire set. 
Among the tested classifiers, Bagging accomplished the highest performance with an ACC of 67.8\% on leave-one-out cross-validation. 
\begin{table}
\caption{Classification performance for bone characterization using individual texture-based classifiers, or their ensembles, as denoted by the block size.}
\label{tab_TCB_BB_Weka_MatLab_BB_LOO}
\begin{center}
\resizebox{.75\textwidth}{!}{
\begin{tabular}{| c | c | c | c | c | c | c |}
\hline
Method  &    Feat. Sel.  &Block Side   & TPR $(\%)$  & TNR $(\%)$  & ACC $(\%)$  & AUC $(\%)$  \\ \hline
NB 		&    CFS-GA	 &400 	&63.2   &64.4   &63.8   &67.3   \\
  		&             	 &100  &38.4   &64.8   &51.7   &54.6  \\
   		&             	 &50   &47.4	&54.7   &52.3   &48.7\\
      	&             	 &25   &46.9	&65.9   &51.7   &55.2\\\hline
BN		&    CFS-GA		 &400 	&66.7   &62.1   &64.4   &70.4   \\
  		&             	 &100  &50.7   &65.4   &59.2   &61.5\\
   		&             	 &50   &37.9	&55.2   &46.6   &50.2\\
      	&             	 &25   &52.4	&54.6   &54.0   &46.0\\\hline
Bagging	&    CFS-GA    	 &400 	&70.1	&65.5	&\textbf{67.8}   &65.0   \\
  		&             	 &100  &44.6   &57.3   &50.6   &52.1\\
   		&             	 &50   &57.8	&58.3   &58.1   &57.4\\
      	&             	 &25   &46.1	&55.1   &51.2   &53.4\\\hline
RF		&    CFS-GA    	 &400 	&67.8	&65.5	&66.7   &68.2  \\
  		&             	 &100  &40.9   &61.6   &51.2   &48.9\\
   		&             	 &50   &45.0	&51.1   &48.3   &50.0\\
      	&             	 &25   &46.8	&53.7   &50.6   &50.7\\\hline
NB 		&    CFS-BF	 &400 	&71.3   &57.5   &64.4   &\textbf{70.9}   \\
  		&             	 &100  &43.9   &59.8   &52.3   &52.0\\
   		&             	 &50   &45.7	&52.3   &50.6   &48.0\\
      	&             	 &25   &48.5	&53.7   &51.7   &49.6\\\hline
BN		&  	 CFS-BF 	 &400 	&64.4   &66.7   &65.5   &69.9   \\
  		&             	 &100  &37.9   &60.9   &49.4   &49.8\\
   		&             	 &50   &40.2	&46.0   &43.1   &44.9\\
      	&             	 &25   &48.8	&53.4   &52.3   &49.9\\\hline
Bagging	&    CFS-BF    	 &400 	&66.6	&67.8	&67.2   &70.5   \\
  		&             	 &100  &50.0   &63.3   &56.9   &52.6\\
   		&             	 &50   &46.7	&53.5   &50.6   &54.1\\
      	&             	 &25   &58.6	&51.0   &54.0   &50.4\\\hline
RF		&    CFS-BF    	 &400 	&60.9	&67.8	&64.4   &68.4   \\
  		&             	 &100  &46.1   &64.7   &55.2   &52.6\\
   		&             	 &50   &48.0	&55.6   &52.3   &52.2\\
      	&             	 &25   &40.9	&44.7   &43.1   &43.4\\\hline
\end{tabular}}
\end{center}
\end{table}

\paragraph{Bag of Keypoints Classifier}
The main parameters that we tuned were the fraction of features to keep for building the vocabulary, the vocabulary size, the penalty coefficient for misclassification of training samples in SVM, and the kernel scale.
The results showed that BoK was able to separate successfully healthy from osteoporotic subjects with an ACC of 99.3\% leave-one-out cross-validation as displayed in Table \ref{tab_TCB_DeeL}. This very high accuracy may be attributed to the extraction of discriminant features from the textured areas. Also, the employed SVM model is known to address data complexity caused by non-linearity and high dimensionality.

\paragraph{Deep Neural Networks} We evaluated the performance of the networks described in Section \ref{sec_DeepLMethods}. We set the learning rates of the convolutional layers to much lower values than the final layers. In this way we largely preserved the pre-trained layer weights at the initial and intermediate convolutional stages. The main parameters that we tuned were the learning rate, learning rate drop, size of mini-batch, and the number of epochs. We utilized grid search and Bayesian optimization search for parameter tuning. Among the deep neural networks, Resnet18 yielded the top ACC of 64.4\% and the top AUC of 67.5\% (Table \ref{tab_TCB_DeeL}).

\begin{table}
\caption{Classification performance for bone characterization using Bag of Keypoints and Deep Learning techniques.}
\label{tab_TCB_DeeL}
\begin{center}
\begin{tabular}{| c | c | c | c | c |}
\hline
Method & TPR $(\%)$ & TNR $(\%)$ & ACC $(\%)$ & AUC $(\%)$ \\\hline
Bag of Keypoints  & 98.6 & 100  & \textbf{99.3} & \textbf{100} \\\hline
AlexNet & 65.5 & 57.5 & 61.5 & 63.1 \\\hline
GoogleNet & 64.4 & 54.0 & 59.2 & 65.6 \\\hline
Resnet18& 80.5 & 48.3 & 64.4 & 67.5 \\\hline
Inceptionv3 & 69.0 & 51.7 & 60.3 & 66.5 \\\hline
\end{tabular}
\end{center}
\end{table}

\paragraph{Conventional SRC}
We then evaluated the performance of the conventional SRC method. We utilized multiple undersampling factors to address convergence to infeasible solutions mostly caused by linearly dependent vectors that yielded different classes. 
In Table \ref{tab_TCB_SRC_[A]_Permuted} we show results from the top performing experiments producing $59.2\%$ classification accuracy for resampling of 1/20, corresponding to feature dimensionality of $400$ using leave-one-out cross-validation. We also applied conventional SRC to the texture feature set produced in Section \ref{sec_feature_selection} and the classification accuracy was 71.7\%.

\paragraph{Integrative Sparse Classification}
Here we report the performance of our block-based ensembles of sparse classifiers. 
We employed block sizes ranging from $100 \times 100$ pixels to $10 \times 10$ pixels to observe the impact of this variable on the classification performance. We repeated these experiments using the BBMAP and BBLL decision functions in this setting.
We show our leave-one-out cross-validation performance in Table \ref{tab_TCB_SRC_[A]_Permuted}.
The experiment with block size $25 \times 25$ pixels that led to 256 classifiers performed the best classification of $100\%$ by the BBMAP and BBLL techniques. These results imply 9.5\% improvement of our method over the traditional SRC method. The block size with $10 \times 10$ also produced 100\% accuracy and 100\% AUC. In addition, we estimated the statistical significance of the differences between the AUC values of BBLL with optimized threshold $\tau^*_{LLS}$ and BBMAP by applying DeLong's statistical test between the ROCs produced by BBMAP and BBLL. The p-values for block sizes of $100 \times 100$, $50 \times 50$, $25 \times 25$ and $10 \times 10$ were $0.47$, $0.66$, $0$ and $0$ respectively, suggesting significant differences for block sizes of $100 \times 100$, $50 \times 50$, $25 \times 25$ and $10 \times 10$. BBLL achieved the top AUC for $25 \times 25$ and $10 \times 10$ block size that was also found to be significantly different from the corresponding BBMAP result. 

\begin{table}
\caption{Classification performance for bone characterization using conventional sparse classifiers and ensembles of block-based sparse classifiers. The block size in the first two rows implies no block decomposition (as in conventional SRC).}
\label{tab_TCB_SRC_[A]_Permuted}
\begin{center}

\begin{tabular}{| p{55pt} | p{55pt} | c | c | c | c |}
\hline
Method & Block Side & TPR $(\%)$  & TNR $(\%)$  & ACC $(\%)$  & AUC $(\%)$	\\\hline
BBMAP & 400 (samp. 1/4)    &55.2  &54.0   &54.6  &58.4 4\\\cline{2-6}
 & 400 (samp. 1/20)   &57.5  &60.9  &59.2  &63.4  \\\cline{2-6}
 & 100  &65.5  &67.8  &66.7  &71.4   \\\cline{2-6}
 & 50 	  &93.1  &81.6  &87.4  &91.3   
  \\\cline{2-6}
 & 25 	  &100   &100   &100   &100   
  \\\cline{2-6}
 & 10 	  &100   &100   &100   &100   
  \\\cline{2-6}
 & Mean$\pm$Std 	  &89.7$\pm$16.4   &87.4$\pm$15.7   &88.5$\pm$15.7   &90.7$\pm$13.5      
  \\\hline
    
BBLL ($\tau^*_{LLS}$ = 0)&  400 (samp. 1/4)   &55.2  &54.0   &54.6  &58.4\\\cline{2-6}
  & 400 (samp. 1/20) &57.5  &60.9  &59.2  &63.4\\\cline{2-6}
 &  100&85.1  &82.8  &83.9  &87.7   
  \\\cline{2-6}        
 &  50	    &98.6  &90.8  &94.8  &97.3   
  \\\cline{2-6}        
 &  25	    &100   &100   &100   &100    
  \\\cline{2-6}        
 &  10 	    &100   &100   &100   &100    
   \\\cline{2-6}        
& Mean$\pm$Std &95.9$\pm$7.2   &93.4$\pm$8.3   &94.7$\pm$7.6   &96.3$\pm$5.8    
  \\\hline
    \end{tabular}
  \end{center}
\end{table}

\section{Discussion}
The cross-validation experiments indicate that BBMAP and BBLL produce better separation than texture-based, BoK, deep neural networks and SRC. Texture-based techniques may achieve moderate classification rates. The feature computation and selection stages are key factors for improving separation accuracy. Smaller block sizes do not improve texture-based classification, because smaller sizes reduce the amount of information represented by the texture descriptors. 

Bag of Keypoints, which shares some similarities with sparse representation methods, showed potential for exceptional results. This method computes patch-based feature vectors that are used for training and testing. SVM classification enables the estimation of non-linear discriminant functions. On the other hand, deep learning methods did not perform very well, mainly because of the limited size of the training set. These methods have potential to outperform their conventional feature-based machine learning counterparts, if we are able to provide a big number of informative labeled samples to train the networks.  

The results in Table \ref{tab_TCB_SRC_[A]_Permuted} suggest that the block-based approach finds more accurate sparse solutions than the conventional SRC approach and improves the classifier performance. A reason for the improved group separation may be that the block-based ensemble technique employs multiple learners of over-complete dictionaries that are more amenable to sparse coding and representation. Between the block-based decision functions, BBLL yielded higher classification rates for larger block sizes, because it accounts for estimation bias. Although both BBMAP and BBLL achieved perfect separation for small blocks, their performance drops when the number of training samples increases and the number of test samples decreases. We expect that effective dictionary learning will help to improve the generalization capability of these methods. 

From the above studies and results, we conclude that machine learning-based methods using digital radiographs, have the potential to assist the non-invasive diagnosis of osteoporosis, or the identification of high fracture risk. A meaningful element of such approaches would be to use the concept of signal sparsity to generate representations of the bone structure.

\bibliographystyle{vancouver}
\bibliography{Chapter_AI_in_Osteoporosis}

\section*{Index Terms}
sparse representation; ensemble classifiers; deep neural networks; texture; computer-aided diagnosis; osteoporosis; fracture risk; radiography

\end{document}